\documentclass{article}

\PassOptionsToPackage{compress}{natbib}
\usepackage[preprint]{arxiv}

\usepackage{zshi}
\usepackage[utf8]{inputenc} 
\usepackage[T1]{fontenc}    
\usepackage{hyperref}       
\usepackage{nicefrac}       
\usepackage{microtype}      
\usepackage{subcaption}
\usepackage{wrapfig}
\usepackage{physics}

\ifdef{\hideNotes}{%
\newcommand{\huan}[1]{}
\newcommand{\zico}[1]{}}
{
\newcommand{\huan}[1]{\textcolor{blue}{Huan: #1}}
\newcommand{\zico}[1]{\textcolor{red}{[Zico: #1]}}
}

\newtoggle{arxiv}
\toggletrue{arxiv}

\def\heuristiclong{Bound Propagation with Shortcuts~(BBPS)}

\title{Neural Network Verification with Branch-and-Bound\\for General Nonlinearities}

\author{
Zhouxing Shi*\textsuperscript{1},
\enskip Qirui Jin*\textsuperscript{2},
\enskip Zico Kolter\textsuperscript{3},
\enskip Suman Jana\textsuperscript{4},
\enskip Cho-Jui Hsieh\textsuperscript{1},
\enskip Huan Zhang\textsuperscript{5}\\
{\normalsize \textsuperscript{1}University of California, Los Angeles \enskip \textsuperscript{2}University of Michigan}\\
{\normalsize
\textsuperscript{3}Carnegie Mellon University
\enskip \textsuperscript{4}Columbia University
\enskip \textsuperscript{5}University of Illinois Urbana-Champaign}\\
{\tt\small z.shi@ucla.edu, qiruijin@umich.edu}\\
{\tt\small zkolter@cs.cmu.edu, suman@cs.columbia.edu, chohsieh@cs.ucla.edu, huan@huan-zhang.com}\\
{\it *Equal contribution}
}

\begin{document}

\maketitle

\begin{abstract}
Branch-and-bound (BaB) is among the most effective techniques for neural network (NN) verification. However, existing works on BaB for NN verification have mostly focused on NNs with piecewise linear activations, especially ReLU networks. In this paper, we develop a general framework, named GenBaB, to conduct BaB on general nonlinearities to verify NNs with general architectures, based on linear bound propagation for NN verification. To decide which neuron to branch, we design a new branching heuristic which leverages linear bounds as shortcuts to efficiently estimate the potential improvement after branching. To decide nontrivial branching points for general nonlinear functions, we propose to pre-optimize branching points, which can be efficiently leveraged during verification with a lookup table. We demonstrate the effectiveness of our GenBaB on verifying a wide range of NNs, including NNs with activation functions such as Sigmoid, Tanh, Sine and GeLU, as well as NNs involving multi-dimensional nonlinear operations such as multiplications in LSTMs and Vision Transformers. Our framework also allows the verification of general nonlinear computation graphs and enables verification applications beyond simple NNs, particularly for AC Optimal Power Flow (ACOPF). 
GenBaB is part of the latest $\alpha,\!\beta$-CROWN\footnote{\url{https://github.com/Verified-Intelligence/alpha-beta-CROWN}}, the winner of the 4th and the 5th International Verification of Neural Networks Competition (VNN-COMP 2023 and 2024). Code for reproducing the experiments is available at \url{https://github.com/shizhouxing/GenBaB}.

\end{abstract}
\section{Introduction}

Neural network (NN) verification aims to formally verify whether a neural network satisfies certain properties, such as safety or robustness properties, prior to its deployment in safety-critical applications. Existing NN verifiers typically compute certified bounds for the output given a pre-defined input region and check the desired properties on the output bounds. As computing exact bounds is NP-complete~\citep{katz2017reluplex}, it becomes crucial to relax the bound computation to improve the efficiency. Bound propagation methods~\citep{wang2018formal,wong2018provable,zhang2018efficient,dvijotham2018dual,henriksen2020efficient,singh2019abstract} have been commonly used, which relax nonlinearities in NNs into linear lower and upper bounds which can be efficiently propagated to finally bound the output of an entire NN. 

To obtain tighter verified bounds, Branch-and-Bound (BaB) has been widely utilized~\citep{bunel2018unified,bunel2020branch,xu2020fast,lu2020neural,de2021improved,wang2021beta,ferrari2021complete} in state-of-the-art NN verifiers, where BaB iteratively \emph{branches} the bounds of intermediate neurons, such that subproblems of verification are created and tighter \emph{bounds} can be computed for each subproblem.  
However, previous works mostly focused on ReLU networks due to the simplicity of ReLU from its piecewise linear nature. Branching a ReLU neuron only requires branching at 0, and it immediately becomes linear in either branch around 0. Conversely, handling NNs with nonlinearities beyond ReLU introduces additional complexity as the convenience of piecewise linearity diminishes. 
It is important for verifying many models with non-ReLU nonlinearities, including:
NNs with non-ReLU activation functions; more complex NNs such as LSTMs~\citep{hochreiter1997long} and Transformers~\citep{vaswani2017attention} which have nonlinearities including multiplication and division beyond activation functions;
applications such as AC Optimal Power Flow (ACOPF)~\citep{guha2019machine} where the verification problem is defined on a computational graph consisting of a NN and also several nonlinear operators encoding the nonlinear constraints to be verified. 
Although some previous works have considered BaB for NNs beyond ReLU networks, e.g., \cite{henriksen2020efficient,wu2022toward} considered BaB on networks with S-shaped activations such as Sigmoid, these works still often specialize in specific and relatively simple types of nonlinearities. A more principled framework for handling general nonlinearities is lacking, leaving ample room for further advancements in verifying non-ReLU NNs.

In this paper, we propose \textbf{GenBaB}, a principled neural network verification framework with BaB for general nonlinearities. To enable BaB for general nonlinearities beyond ReLU, we first formulate a general BaB framework, and we introduce general branching points, where we may branch at points other than 0 for nonlinear functions, which is needed when the nonlinearity is not piecewise linear around 0. We then propose a new branching heuristic named ``\heuristiclong'' for branching general nonlinearities, which carefully leverages the linear bounds from bound propagation as shortcuts to efficiently and effectively estimate the bound improvement from branching a neuron. Moreover, we propose to decide nontrivial branching points by pre-optimizing branching points, according to the tightness of the resulted linear relaxation, and we save the optimized branching points into a lookup table to be efficiently used when verifying an entire NN with different data instances. 

We demonstrate the effectiveness of our GenBaB on a variety of networks, including feedforward networks with Sigmoid, Tanh, Sine, or GeLU activations, as well as LSTMs and Vision Transformers (ViTs). These models involve various nonlinearities including S-shaped activations, periodic trigonometric functions, and also multiplication and division which are multi-dimensional nonlinear operations beyond activation functions. We also enable verification on models for the AC Optimal Power Flow (ACOPF) application~\citep{guha2019machine}. GenBaB is generally effective and outperforms existing baselines. The improvement from GenBaB is particularly significant for models involving functions with stronger nonlinearity. For example, on a $4\times 100$ network with the Sine activation, GenBaB improves the verification from 4\% to 60\% instances verified (NNs with the Sine activation have been proposed for neural representations and neural rendering in \citet{sitzmann2020implicit}). 


\section{Background}
\label{sec:bg}

\textbf{The NN verification problem.}
Let $f: \sR^{d}\rightarrow \sR^{K}$ be a neural network taking input $\rvx\in\sR^{d}$ and outputting $f(\rvx)\in\sR^{K}$. Suppose $\gC$ is the input region to be verified, and $s:\sR^{K}\rightarrow \sR$ is an output specification function, $h:\sR^{d}\mapsto\sR$ is the function that combines the NN and the output specification as $h(\rvx)=s(f(\rvx))$.
NN verification can typically be formulated as verifying if $h(\rvx)>0, \forall \rvx\in\gC$ provably holds.
A commonly adopted special case is robustness verification given a small input region, where $f(\rvx)$ is a $K$-way classifier and $h(\rvx)\coloneqq \min_{i\neq c}\{ f_c(\rvx) - f_i (\rvx)\}$ checks the worst-case margin between the ground-truth class $c$ and any other class $i$.
The input region is often taken as a small $\ell_\infty$-ball with radius $\eps$ around a data point $\rvx_0$, i.e., $\gC\coloneqq \{ \rvx \mid \|\rvx-\rvx_0\|_\infty\!\leq\!\eps \}$. This is a succinct and useful problem for provably verifying the robustness properties of a model and also for benchmarking NN verifiers, although there are other NN verification problems beyond robustness~\citep{brix2023fourth}. We mainly focus on this setting for its simplicity following prior works.

\textbf{Linear bound propagation}.
We develop our GenBaB based on linear bound propagation~\citep{zhang2018efficient,xu2020automatic} for computing the verified bounds of each subproblem during the BaB.
Linear bound propagation can lower bound $h(\rvx)$ by propagating linear bounds w.r.t. the output of one or more intermediate layers as
$h(\rvx)\geq  \sum\nolimits_i \rmA_i \hat{\rvx}_i + \rvc~(\forall\rvx\in\gC)$,
where $ \hat{\rvx}_i~(i\leq n)$ is the output of intermediate layer $i$ in the network $f(\rvx)$ with $n$ layers, $\rmA_i$ are the coefficients w.r.t. layer $i$, and $\rvc$ is a bias term.
In the beginning, the linear bound is simply $h(\rvx)\geq \rmI \cdot h(\rvx) + \vzero $ which is actually an equality. In the bound propagation, $ \rmA_i \hat{\rvx}_i $ 
is recursively substituted by the linear bound of $\hat{\rvx}_i$ w.r.t its input. For simplicity, suppose layer $i-1$ is the input to layer $i$ and $\hat{\rvx}_i=h_i(\hat{\rvx}_{i-1})$, where $h_i(\cdot)$ is the computation for layer $i$. And suppose we have the  linear bounds of $\hat{\rvx}_i$ w.r.t its input $\hat{\rvx}_{i-1}$ as:
\begin{equation}
\ul{\rva}_i \hat{\rvx}_{i-1} + \ul{\rvb}_i
\leq \hat{\rvx}_i=h_i(\hat{\rvx}_{i-1})
\leq \ol{\rva}_i \hat{\rvx}_{i-1} + \ol{\rvb}_i,
\label{eq:relaxation}
\end{equation}
with parameters $\ul{\rva}_i,\ul{\rvb}_i,\ol{\rva}_i,\ol{\rvb}_i$ for the linear bounds, and ``$\leq$'' holds elementwise.
Then $\rmA_i\hat{\rvx}_i$ can be substituted and lower bounded by
$\rmA_i\hat{\rvx}_i\geq
\rmA_{i-1}
\hat{\rvx}_{i-1}
+\big(\rmA_{i,+} \ul{\rvb}_i
+\rmA_{i,-} \ol{\rvb}_i\big)$,
where $\rmA_{i-1}=\rmA_{i,+} \ul{\rva}_i
+\rmA_{i,-} \ol{\rva}_i$,
(``+'' and ``-'' in the subscripts denote taking positive and negative elements, respectively).
In this way, the linear bounds are propagated from layer $i$ to layer $i-1$.
Ultimately, the linear bounds can be propagated to the input of the network $\rvx$ as $ h(\rvx)\geq \rmA_0 \rvx + \rvc~(\rmA_0\in\sR^{1\times d})$, where the input can be viewed as the 0-th layer.
Depending on $\gC$, this linear bound can be concretized into a lower bound without $\rvx$. 
We focus on settings where $\gC$ is an $\ell_\infty$-ball as mentioned above, and thereby we have: 
\begin{equation}
\forall\|\rvx-\rvx_0\|_\infty\leq\eps,\quad\rmA_0\rvx+\rvc\geq \rmA_0\rvx_0 - \eps \|\rmA_0\|_{1} + \rvc.
\label{eq:concretize}
\end{equation}

To obtain \eqref{eq:relaxation}, if $h_i$ is a linear operator, we simply have
$\ul{\rva}_i \hat{\rvx}_{i-1} + \ul{\rvb}_i= \ol{\rva}_i \hat{\rvx}_{i-1} + \ol{\rvb}_i=h_i(\hat{\rvx}_{i-1})$ which is $h_i$ itself.
Otherwise, linear relaxation is used, which relaxes a nonlinearity and  bound the nonlinearity by linear functions. An intermediate bound on $\hat{\rvx}_{i-1}$ as $\rvl_{i-1}\leq \hat{\rvx}_{i-1}\leq\rvu_{i-1}$ is usually required for the relaxation, which can be obtained by treating the intermediate layer as the output of a network and running additional bound propagation.

\section{Method}
\label{sec:method}

\subsection{Overall Framework}
\label{sec:framework}

\textbf{Notations.} Although in \Cref{sec:bg}, we considered a feedforward NN for simplicity, linear bound propagation has been generalized to NNs with general architectures and general computational graphs~\citep{xu2020automatic}. In our work, we also consider a general computational graph $h(\rvx)$ for input region $\rvx\in\gC$. Instead of a feedforward network with $n$ \emph{layers} in \Cref{sec:bg}, we consider a computational graph with $n$ \emph{nodes}, where each node $i$ computes some function $ h_i(\cdot)$ which may either correspond to a linear layer in the NN or a nonlinearity. We use $\hat{\rvx}_i$ to denote the output of node $i$, which may contain many neurons, and we use  $\hat{\rvx}_{i,j}$ to denote the output of the $j$-th neuron in node $i$.  Intermediate bounds of node $i$ may be needed to relax and bound $h_i(\cdot)$, and we use $ \rvl_{i,j}, \rvu_{i,j}$ to denote the intermediate lower and upper bound respectively. We use $\rvl$ and $\rvu$ to denote all the intermediate lower bounds and upper bounds, respectively, for the entire computational graph.

\paragraph{Overview of GenBaB.}
\Cref{fig:framework} illustrates our GenBaB framework.
Our GenBaB is a general branch-and-bound framework to handle NNs with general nonlinearities, for NN verification with linear bound propagation. 
Note that our contributions focus on the \emph{branching} part for general nonlinearities, while \emph{bounding} for individual subdomains during BaB follows existing linear bound propagation which has supported general models~\citep{xu2020automatic}.  

We conduct an initial verification using linear bound propagation before entering BaB. We proceed to BaB only if the initial verification is not sufficient for a successful verification, and we aim to use BaB to enhance the verification for such hard cases. 
In our BaB, we branch the intermediate bounds of neurons connected to general nonlinearities. 
We maintain a dynamic pool of intermediate bound domains,
$\gD=\{ (\rvl^{(i)},\rvu^{(i)})\}_{i=1}^m$,
where 
each domain $(\rvl^{(i)},\rvu^{(i)})~(1\leq i\leq m)$ denotes the intermediate bounds of a subproblem in the BaB, $m=|\gD|$ is the number of current domains, and initially we have $\gD=\{ (\rvl,\rvu)\}$ with the intermediate bounds from the initial verification.
Then in each iteration of BaB, we pop a domain from $\gD$, and we select a neuron to branch and a branching point between the intermediate bounds of the selected neuron. 
To support general nonlinearities, we formulate a new and general branching framework in \Cref{sec:general_branching}, where we introduce general branching points, in contrast to branching ReLU at 0 only, and we also support more complicated networks architectures where a nonlinearity can involve multiple input nodes or output nodes.
To decide nontrivial branching points, in \Cref{sec:branching_points}, we propose to pre-optimize the branching points, which aims to produce the tightest linear relaxation after taking the optimized branching point. 
And in order to decide which neuron we choose to branch, we propose a new branching heuristic in \Cref{sec:heuristic} to estimate the potential improvement for each choice of a branched neuron, where we carefully leverage linear bounds as an efficient shortcut for a more precise estimation. 

Each branching step generates new subdomains. For the new subdomains, we update $\rvl,\rvu$ for the branched neurons according to the branching points, and the branching decision is also encoded into the bound propagation as additional constraints by Lagrange multipliers following \citet{wang2021beta}.
For each new subdomain, given updated $\rvl,\rvu$, we use $V(h,\gC,\rvl,\rvu)$ to denote a new verified bound computed with new intermediate bounds $\rvl,\rvu$. Subdomains with $ V(h, \gC,\rvl, \rvu)>0$ are verified and discarded, otherwise they are added to $\gD$ for further branching. We repeat the process until no domain is left in $\gD$ and the verification succeeds, or when the timeout is reached and the verification fails.
In the implementation, our BaB is batched where many domains are handled in parallel on a GPU with the batch size dynamically tuned to fit the GPU memory.

\subsection{Branching for General Nonlinearities}
\label{sec:general_branching}

\begin{figure}[t]
\centering
\includegraphics[width=.88\textwidth,trim={0 50pt 0 25pt},clip]{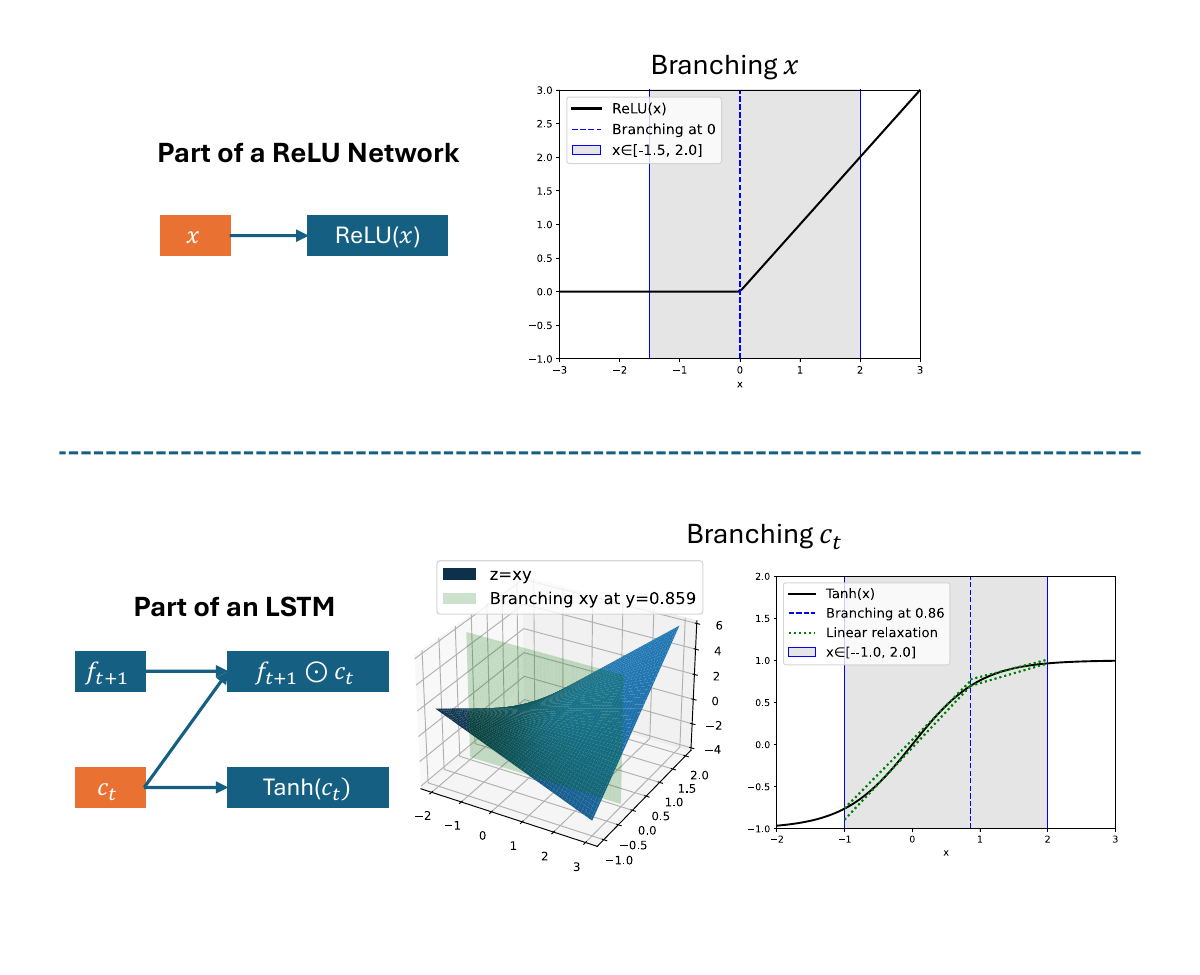}
\caption{
Illustration of the more complicated nature of branching for general nonlinearities (branching a ReLU activation v.s. branching for nonlinearities in an LSTM). 
Notations for part of an LSTM follows PyTorch's documentation~(\url{https://pytorch.org/docs/stable/generated/torch.nn.LSTM.html}). Nodes in orange are being branched.
For general nonlinearities, branching points can be non-zero (0.86 in the LSTM example here), a nonlinearity can take multiple input nodes ($f_{t+1}\odot c_t$ here), and a node can also be followed by multiple nonlinearities ($c_t$ is followed by a multiplication and also Tanh, and branching $c_t$ affects both two nonlinearities). 
}
\label{fig:branching}  
\end{figure}

As illustrated in \Cref{fig:branching}, branching for general nonlinearities on general computational graphs is more complicated, in contrast to BaB for ReLU networks.
For general nonlinearities, we need to consider branching at points other than 0. 
In addition, unlike typical activation functions, some nonlinearities may take more than one inputs and thereby have multiple input nodes that can be branched, such as multiplication in LSTM (``$f_{t+1}\odot c_t$'' in \Cref{fig:branching}) or Transformers~\citep{hochreiter1997long,vaswani2017attention}. On general computational graphs, a node can also be followed by multiple nonlinearities, as appeared in LSTMs (such as ``$c_t$'' in \Cref{fig:branching}), and then branching the intermediate bounds of this node can affect multiple nonlinearities.

To resolve these challenges, we propose a more general formulation for branching on general computational graphs with general nonlinearities. Each time, we consider branching the intermediate bounds of a neuron $j$ in a node $i$, namely $ [\rvl_{i,j}, \rvu_{i,j}]$, if node $i$ is the input of some nonlinearity. We consider branching the concerned neuron into 2 branches with a nontrivial branching point $\rvp_{i,j}$, as
$[\rvl_{i,j}, \rvu_{i,j}] \rightarrow
[\rvl_{i,j},\rvp_{i,j}],~[\rvp_{i,j},\rvu_{i,j}]$.
Here we consider branching from the perspective of each node $i$ which is the input to at least one nonlinearity and decide if we branch the intermediate bounds $[\rvl_i,\rvu_i]$ of this node.
This consideration allows us to conveniently support nonlinearities with multiple input nodes or multiple nonlinearities sharing an input node.
On the contrary, if we consider branching from the perspective of each nonlinearity, the considered nonlinearity may share some input node with another nonlinearity and thus other nonlinearities can also be affected.

\subsection{Where to Branch? New Considerations for General Nonlinear Functions}
\label{sec:branching_points}

\begin{figure*}[ht]
    \centering
    \begin{subfigure}[t]{0.31\textwidth}
    \centering
    \includegraphics[width=\textwidth]{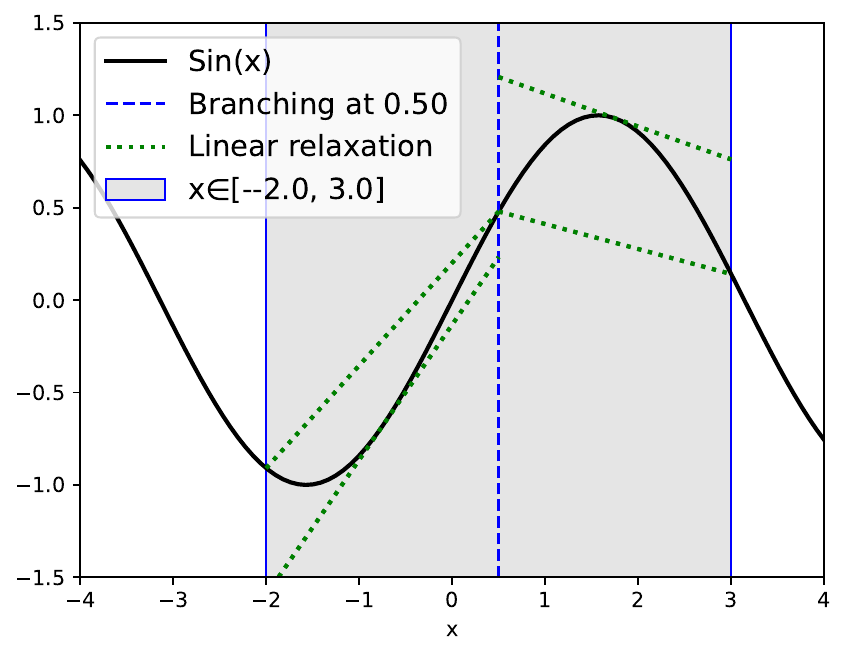}
    \caption{Branching a Sine activation in the middle.}
    \label{fig:sin_middle}
    \end{subfigure}
    \hfill
    \begin{subfigure}[t]{0.31\textwidth}
    \centering
    \includegraphics[width=\textwidth]{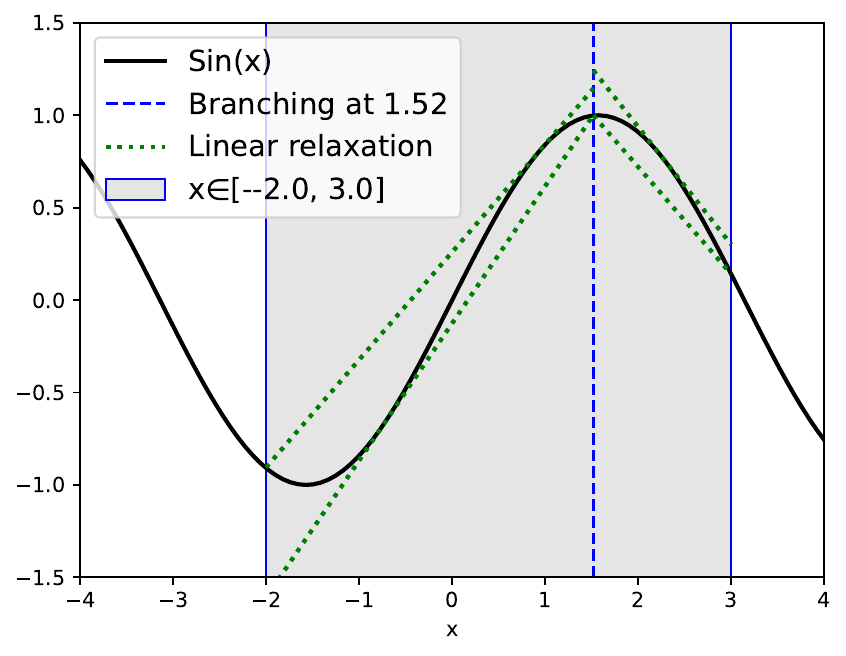}
    \caption{Branching a Sine at our pre-optimized branching point.}
    \label{fig:sin_opt}
    \end{subfigure}
    \hfill
    \begin{subfigure}[t]{0.31\textwidth}
    \centering
    \includegraphics[width=\textwidth]{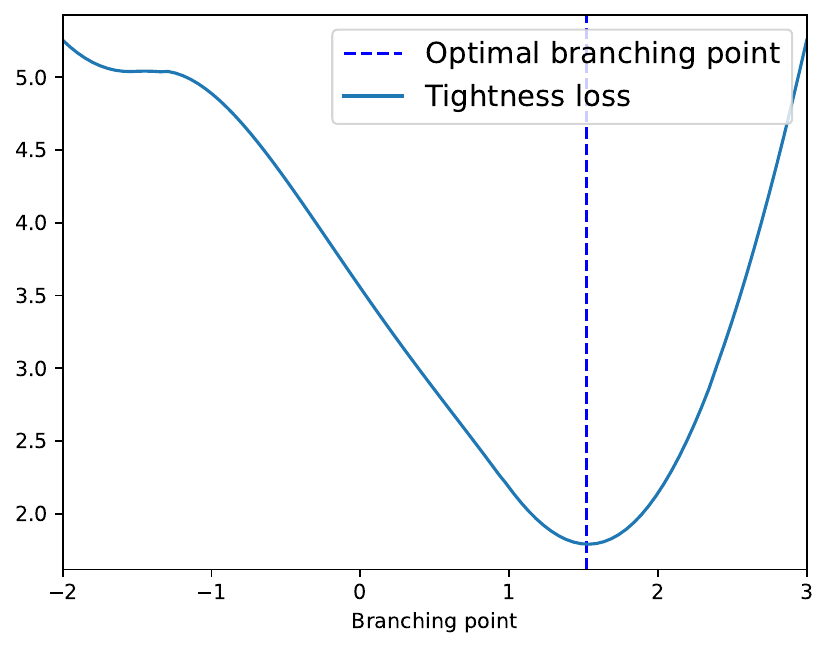}
    \caption{The tightness loss defined in \eqref{eq:opt_bp} for different branching points.}
    \label{fig:sin_loss}
    \end{subfigure}
    \vspace{-5pt}
    \caption{Illustration of branching the intermediate bounds of a neuron connected to the Sine activation~\citep{sitzmann2020implicit}. 
    }
    \label{fig:branching_points}
    \end{figure*}

The more complex nature of general nonlinear functions also brings flexibility on choosing branching points, compared to the ReLU activation where only branching at 0 is reasonable. 
A straightforward way is to branch in the middle between the intermediate lower and upper bounds, as shown in \Cref{fig:sin_middle}. However, this can be suboptimal for many nonlinear functions.
Intuitively, as tighter linear relaxation can often lead to tighter verified bounds~\citep{lyu2019fastened,xu2020fast}, we aim to choose a branching point such that the linear relaxation for both sides after the branching can be as tight as possible. 
Therefore, we propose to \emph{pre-optimize branching points} for each case of nonlinearity in the model, before actually running BaB on different data instances. 
We enumerate all pairs of possible intermediate bounds within a certain range with a step size, where we set a small step size which defines the gap between the adjacent enumerated intermediate bounds. And we save the optimized branching points into a lookup table. During verification, for each pair of intermediate bounds we actually encounter, we efficiently query the lookup table and take the branching point for the closest intermediate bound pair in the lookup table (if no valid branching point is obtained from the lookup table, we try branching in the middle instead as a backup).
An example of pre-optimized branching points is shown in \Cref{fig:sin_opt}. We only need to pre-optimize branching points once for each new model, and the produced lookup table can be used on an arbitrary number of data instances, and thus the time cost of the pre-optimization is negligible for the overall verification. 

We now formulate the objective of the pre-optimization.
For simplicity here, we mainly assume that we have a unary nonlinear function $q(x)$, although our method supports functions with any number of inputs in practice.
Suppose the input intermediate bounds for $q(x)$ is $l\leq x\leq u$, we aim to find a branching point $p=P(l,u)$ such that the overall tightness of the linear relaxation for input range $[l,p]$ and $[p,u]$, respectively, is the best.
Suppose the linear relaxation for input range $[l,p]$ is $\ul{a}_1 x + \ul{b}_1 \leq q(x) \leq \ol{a}_1 x + \ol{b}_1 $, and similarly   $\ul{a}_2 x + \ul{b}_2 \leq q(x) \leq \ol{a}_2 x + \ol{b}_2 $ for input range $[p,u]$. 
Following previous works such as \citet{shi2020robustness}, we use the integral of the gap between the lower linear relaxation and the upper linear relaxation to measure the tightness (the linear relaxation is considered as tighter when the gap is smaller).
We define it as a \emph{tightness loss} $P(q(x),l,u,p)$ for nonlinearity $q(x)$ with input range $[l,u]$ and branching point $p$:
\begin{equation}
P(q(x),l,u,p)=
\int_{l}^p
\bigg(\big(
\ol{a}_1 x + \ol{b}_1
\big)
- \big(
\ul{a}_1 x + \ul{b}_1
\big)\bigg)\dd{x}
+
\int_{p}^{u}
\bigg(\big(
\ol{a}_2 x + \ol{b}_2
\big)
- \big(
\ul{a}_2 x + \ul{b}_2
\big)\bigg)\dd{x},
\label{eq:opt_bp}
\end{equation}
where the parameters for the linear relaxation ($\ul{a}_1,\ol{a}_1,\ul{b}_1,\ol{b}_1,\ul{a}_2,\ol{a}_2,\ul{b}_2,\ol{b}_2$) all depend on $p$.
We take the best branching point $p~(l<p<u)$ which minimizes $P(q(x), l,u,p)$.
\Cref{fig:sin_loss} plots the tightness loss for the Sine activation.
This problem can be solved by gradient descent, or an enumeration over a number of potential branching points if the nonlinear function only has one or two inputs.

Moreover, we also support a generalized version of \eqref{eq:opt_bp} for nonlinear functions with multiple inputs (such as multiplication involving two inputs), where we use a multiple integral to measure the tightness for multi-dimensional nonlinearities.
And when a branched node has multiple nonlinear output nodes, we take the sum for multiple nonlinearities as $\sum_{q \in \mathcal{Q}} P(q(x), l, u,p)$, where $\mathcal{Q}$ is the set of output nonlinearities. As such, our pre-optimized branching points support general computational graphs.

\subsection{Which Neuron to Branch? A New Branching Heuristic}
\label{sec:heuristic}

Since a NN usually contains many neurons where branching can potentially occur, typically a branching heuristic is used to efficiently decide a neuron to branch, so that the time cost of each BaB iteration is moderate to allow more BaB iterations within the time budget. The branching heuristic is essentially a scoring function for estimating the new verified bound after branching at each neuron, in order to choose a good neuron which potentially leads to a good improvement after the branching. 
We propose a new branching heuristic to support general nonlinearities. 

Specifically, we design a function $\tilde{V}(\rvl,\rvu,i,j,k,\rvp_{i,j})$ which estimates the new bound of the $k$-th ($1\!\leq\! k\!\leq\!2$) branch, after branching neuron $j$ in node $i$ using branching points $\rvp_{i,j}$.
We use $B(\rvl,\rvu,i,j,k,\rvp_{i,j})$ to denote the updated intermediate bounds after this branching, and essentially we aim to use $\tilde{V}(\rvl,\rvu,i,j,k,\rvp_{i,j})$ to efficiently estimate $V(h, \gC,B(\rvl,\rvu,i,j,k,\rvp_{i,j}))$ which is the actual verified bound after the branching, but it is too costly to directly compute an actual verified bound for each branching option.

Suppose we consider branching a neuron $j$ in node $i$ and we aim to estimate $V(\cdot)$ for each branch $k$.
In the linear bound propagation, when the bounds are propagated to node $i$, we have:
\begin{align}
h(\rvx)\geq &~\rmA_{i,j}^{(k)} \hat{\rvx}_{i,j} +\rvc^{(k)}
\geq V (h, \gC, B(\rvl,i,j,k,\rvp_{i,j})),
\label{eq:heuristic_income}
\end{align}
where we use $\rmA_{i,j}^{(k)}$ and $\rvc^{(k)}$ to denote the parameters in the linear bounds for the $k$-th branch, and here $\rvc^{(k)}$ a bias term accumulated on all the neurons.
Since we do not update the intermediate bounds except for the branched neurons during BaB for efficiency following \citet{wang2021beta}, branching a neuron in node $i$ only affects the linear relaxation of nonlinear nodes immediately after node $i$ (i.e., output nodes of $i$).
Therefore, $\rmA_{i,j}^{(k)}$ and $\rvc^{(k)}$ can be computed by only propagating the linear bounds from the output nodes of $i$, using previously stored linear bounds, rather than from the final output of $h(\rvx)$.

For a more efficient estimation, instead of  propagating the linear bounds towards the input of the network step by step, we propose a new branching heuristic named \emph{\heuristiclong}, where we use a shortcut to directly propagate the bounds to the input.
Specifically, we save the linear bounds of all the potentially branched intermediate nodes during the initial verification. For every neuron $j$ in intermediate node $i$, we record:
\begin{equation}
\forall\rvx\in\gC,\quad
\ul{\hat{\rmA}}_{ij}\rvx+\ul{\hat{\rvc}}_{ij}
\leq \hat{\rvx}_{ij}
\leq \ol{\hat{\rmA}}_{ij}\rvx+\ol{\hat{\rvc}}_{ij},
\label{eq:saved_A}
\end{equation}
where $\ul{\hat{\rmA}}_{ij},\ul{\hat{\rvc}}_{ij} ,\ol{\hat{\rmA}}_{ij},\ol{\hat{\rvc}}_{ij}$ are parameters for the linear bounds. These are obtained when linear bound propagation is used for computing the intermediate bounds $[\rvl_{i,j},\rvu_{i,j}]$ and the linear bounds are propagated to the input $\rvx$. We then use \eqref{eq:saved_A} to compute a lower bound for $\rmA_{i,j}^{(k)} \hat{\rvx}_{i,j} +\rvc^{(k)}$:
\begin{align}
\rmA_{i,j}^{(k)} \hat{\rvx}_{i,j} +\rvc^{(k)}
\geq
&~(\rmA_{i,j,+}^{(k)}\ul{\hat{\rmA}}_{ij}
+\rmA_{i,j,-}^{(k)}\ol{\hat{\rmA}}_{ij})\rvx
+\rmA_{i,j,+}^{(k)}\ul{\hat{\rvc}}_{ij}
+\rmA_{i,j,-}^{(k)}\ol{\hat{\rvc}}_{ij}
+\rvc^{(k)}.
\label{eq:apply_shortcut}
\end{align}
The right-hand-side can be concretized by \eqref{eq:concretize} to serve as an approximation for $V(\cdot)$ after the branching. In this way, the linear bounds are directly propagated from node $i$ to input $\rvx$ and concretized using a shortcut.
We thereby take the concretized bound as $\tilde{V}(\rvl,\rvu,i,j,k,\rvp_{i,j})$ for our BBPS heuristic score.

This computation is efficient, and it does not affect the time complexity of BaB as the time complexity is mainly dominated by the bound computation after each branching.
Our branching heuristic is also generally formulated. We leverage updates on the linear relaxation of any nonlinearity, and general branching points and general number of inputs nodes are supported when we update the linear relaxation. Node $i$ can also have multiple nonlinear output nodes, as we accumulate the linear bounds propagated from all the output nodes to produce \eqref{eq:heuristic_income}. 

\paragraph{Comparison to branching heuristics in previous works.}
Existing branching heuristics from previous works~\citep{bunel2018unified,bunel2020branch,lu2020neural,de2021improved} are more restrictive, as they mostly focused on branching ReLU neurons with a fixed branching point (0 for ReLU) and their heuristic is specifically formulated for ReLU, unlike our general formulation above.
Even if we directly generalize their branching heuristic to support general nonlinearities, we also empirically find they are often not precise enough for general nonlinearities due to their more aggressive approximation. In the existing BaBSR heuristic originally for ReLU networks~\citep{bunel2020branch}, they essentially propagate the bounds only to the node before the branched one with an early stop, and they then ignore the coefficients ($\rmA_{i-1,j}^{(k)}$ for a feedforward NN) without propagating further. In contrast, in our BBPS heuristic, we carefully utilize a shortcut to propagate the bounds to the input as \eqref{eq:apply_shortcut} rather than discard linear terms early. Therefore, we expect our BBPS heuristic to be more precise and effective.

\section{Experiments}
\label{sec:experiments}

\subsection{Settings}

Implementation and additional experimental details are provided in \Cref{ap:details}.

\paragraph{Models and Data.}
We focus on verifying NNs with nonlinearities beyond ReLU, and we experiment on models with various nonlinearities as shown in \Cref{tab:models}. We mainly consider the commonly used $\ell_\infty$ robustness verification specification on image classification. We use the term \emph{instance} to refer to a data example along with the corresponding verification specification.
We adopt some MNIST~\citep{mnist} models with Sigmoid and Tanh activation functions from previous works~\citep{singh2019beyond,singh2019abstract,muller2022prima}, along with their data instances. 
Besides, to test our method on more models with various nonlinearities using a consistent training setting for all the models, we train many new models with various nonlinearities on CIFAR-10~\citep{cifar} by PGD adversarial training~\citep{madry2017towards}, using an $\ell_\infty$ perturbation with $\eps=1/255$ in both training and verification. 
The models we train on CIFAR-10 include models with Sigmoid, Tanh, Sine, and GeLU activation functions, respectively, as well as LSTM~\citep{hochreiter1997long} and ViT~\citep{dosovitskiy2020image}. 
We adopt PGD adversarial training, because NNs trained without robust training are known to be highly vulnerable to tiny adversarial perturbations~\citep{szegedy2013intriguing,goodfellow2014explaining} and formal verification is not possible unless $\eps$ is much smaller.
For these CIFAR-10 models, we first run vanilla CROWN~\citep{zhang2019towards,xu2020automatic} (linear bound propagation without optimized linear relaxation~\citep{xu2020fast,lyu2019fastened} or BaB~\citep{xu2020fast,wang2021beta}), to remove instances which are too easy where vanilla CROWN already succeeds. We also remove instances where PGD attack succeeds, as such instances are impossible to verify. 
We only retain the first 100 instances if there are more instances left. We set a timeout of 300 seconds for our BaB in all these experiments. 
In addition, we adopt an NN verification benchmark for verifying properties in the Machine Learning for AC Optimal Power Flow (ML4ACOPF) problem~\citep{guha2019machine}\footnote{Benchmark: \url{https://github.com/AI4OPT/ml4acopf_benchmark}.} which is beyond robustness verification. 
In the Appendix, we have results on additional models:
a ResNet model~\citep{he2016deep} in \Cref{ap:resnet};
models with larger $\eps=2/255$ and $\eps=8/255$ in \Cref{ap:large_eps};
and a ReLU model in \Cref{ap:relu}, included for completeness.
\begin{wraptable}{r}{.5\textwidth}
\centering
\caption{List of models with various nonlinearities in our experiments.}
\adjustbox{max width=.5\textwidth}{
\begin{tabular}{ll}
\toprule
Model & Nonlinearities in the model\\
\midrule
Feedforward & $\operatorname{sigmoid}, \tanh, \sin$, GeLU\\
LSTM & $\operatorname{sigmoid}$, $\tanh$, $xy$\\
ViT with ReLU & ReLU, $xy$, $x/y$, $x^2$, $\sqrt{x}$, $\exp(x)$ \\
ML4ACOPF & ReLU, $\operatorname{sigmoid}$, $\sin$, $xy$, $x^2$\\
\bottomrule
\end{tabular}}
\label{tab:models}
\end{wraptable}

\paragraph{Baselines.}
We compare our GenBaB with the previous \abcrown which did not support BaB on non-ReLU nonlinearities. 
We also compare with several other baselines, including DeepPoly~\citep{singh2019abstract}, PRIMA~\citep{muller2022prima}, VeriNet~\citep{henriksen2020efficient},  PROVER~\citep{ryou2021scalable}, DeepT~\citep{bonaert2021fast}, \citet{wu2022toward}, \citet{wei2023convex}, on the models they support, respectively. Among these baselines, only VeriNet and \citet{wu2022toward} support BaB on Sigmoid or Tanh models, and none of the baseline supports BaB on general nonlinearities. 
While the original BaBSR heuristic in \citet{bunel2020branch} only supported ReLU networks, we also implemented a generalized version of BaBSR for nonlinearities beyond ReLU for an empirical comparison in \Cref{tab:cifar}, based on the difference in treating the linear term discussed in \Cref{sec:heuristic}.

\subsection{Main Results}

\begin{table*}[ht]
\centering
\caption{Number of verified instances out of the first 100 test examples on MNIST for several Sigmoid networks and Tanh networks along with their $\eps$. The settings are the same as those in PRIMA~\citep{muller2022prima}. ``$L\times W$'' in the network names denote a fully-connected NN with $L$ layers and $W$ hidden neurons in each layer. The upper bounds in the last row are computed by PGD attack~\citep{madry2017towards}, as a sound verification should not verify instances where PGD can successfully find counterexamples.
}

\adjustbox{max width=.95\textwidth}{
\begin{tabular}{l|cccc|cccc}
\toprule
\multirow{3}{*}{Method} & \multicolumn{4}{c|}{Sigmoid Networks} & \multicolumn{4}{c}{Tanh Networks}\\
& 6$\times$100 & 6$\times$200 & 9$\times$100 & ConvSmall & 6$\times$100 & 6$\times$200 & 9$\times$100 & ConvSmall\\
& $\eps\!=\!0.015$ & $\eps\!=\!0.012$ & $\eps\!=\!0.015$ & $\eps\!=\!0.014$ & $\eps\!=\!0.006$ & $\eps\!=\!0.002$ & $\eps\!=\!0.006$ & $\eps\!=\!0.005$\\
\midrule
DeepPoly$^\text{a}$$^\text{b}$ & 30 & 43 & 38 & 30 & 38 & 39 & 18 & 16\\
PRIMA$^\text{a}$ & 53 & 73 & 56 & 51 & 61 & 68 & 52 & 30\\
VeriNet$^\text{c}$ & 65 & 81 & 56 & - & 31 & 30 & 16 & - \\
Marabou~\citep{wu2022toward}$^?$ 
& {\color{gray}65} & {\color{gray}75} & {\color{gray}96$^?$} & {\color{gray}63} & - & - & - & -\\
Vanilla CROWN$^\text{b}$ & 53 & 63 & 49 & 65 & 18 & 24 & 44 & 55\\
\abcrown (w/o BaB) & 62 & 81 & \textbf{62} & 84 & \textbf{65} & 72 & 58 & 69\\
\textbf{GenBaB (ours)} & \textbf{71} & \textbf{83} & \textbf{62} & \textbf{92} & \textbf{65} & \textbf{78} & \textbf{59} & \textbf{75}\\
\midrule
Upper bound & 93 & 99 & 92 & 97 & 94 & 97 & 96 & 98\\
\bottomrule
\end{tabular}}

\begin{minipage}{\linewidth}\footnotesize
$^\text{a}$Results for DeepPoly and PRIMA are directly from \citet{muller2022prima}.\\
$^\text{b}$While DeepPoly and CROWN are thought to be equivalent on ReLU networks~\citep{muller2022prima}, these two works adopt different relaxation for Sigmoid and Tanh, which results in different results here.\\
$^\text{c}$Results for VeriNet are obtained by running the tool 
(\url{https://github.com/vas-group-imperial/VeriNet}) by ourselves.
VeriNet depends on the FICO Xpress commercial solver which requires a license for models that are relatively large. FICO Xpress declined the request we submitted for an academic license due to the lack of a course tutor. Thus, results on ConvSmall models are not available.\\
$^{?}$We found that the result \citet{wu2022toward} reported on the Sigmoid $9\times 100$ model exceeds the upper bound by PGD attack ($96>92$), and thus the result tends to be not fully valid (also reported in \citet{zhou2024testing}). 
\end{minipage}
\label{tab:mnist_eran}
\end{table*}

\paragraph{Experiments on Sigmoid and Tanh networks for MNIST.}
We first experiment on Sigmoid networks and Tanh networks for MNIST and show the results in \Cref{tab:mnist_eran}.
On 6 out of the 8 models, our GenBaB is able to verify more instances over \abcrown without BaB and also outperforms all the non-CROWN baselines. We find that improving on Sigmoid $9\times 100$ and Tanh $6\times 100$ networks by BaB is harder, as the initial bounds are typically too loose on the unverifiable instances before BaB, possibly due to these models being trained without robustness consideration.

\begin{table*}[t]
\centering
\caption{Number of verified instances out of 100 filtered instances on CIFAR-10 with $\eps=1/255$ for feedforward NNs with various activation functions.
The last three rows contain results for the ablation study, where ``Base BaB'' does not use our BBPS heuristic or pre-optimized branching points, but it uses a generalized BaBSR heuristic~\citep{bunel2020branch} and always branches intermediate bounds in the middle.
}
\adjustbox{max width=\textwidth}{
\begin{tabular}{lccccccccccccc}
\toprule
\multirow{2}{*}{Method} & \multicolumn{4}{c}{Sigmoid Networks} & \multicolumn{2}{c}{Tanh Networks} & \multicolumn{3}{c}{Sine Networks} & \multicolumn{3}{c}{GeLU Networks}\\
& 4$\times$100 & 4$\times$500
& 6$\times$100 & 6$\times 200$
& 4$\times$100 & 6$\times 100$
& 4$\times$100 & 4$\times 200$
& 4$\times$500 & 4$\times 100$
& 4$\times$200 & 4$\times 500$\\
\midrule
PRIMA$^\text{a}$ & 0 & 0 & 0 & 0 & 0 & 0 & - & - & - & - & - & - \\
Vanilla CROWN$^\text{b}$ &  0 & 0 & 0 & 0 & 0 & 0 & 0 & 0 & 0 & 0 & 0 & 0\\
\abcrown w/o BaB$^\text{c}$ & 28 & 16 & 43 & 39 & 25 & 6 & 4 & 2 & 4 & 44 & 33 & 27 \\
\textbf{GenBaB (ours)} & 
\textbf{58} & \textbf{24} & \textbf{64} & \textbf{50} 
& \textbf{49} & \textbf{10} 
& \textbf{60} & \textbf{35} & \textbf{22} 
& \textbf{82} & \textbf{65} & \textbf{39}\\
\hline
\multicolumn{13}{c}{Ablation Studies}\\
\hline
Base BaB & 
34 & 19 & 44 & 41 
& 34 & 8 
& 9 & 8 & 7
& 64 & 54 & 39\\
+ BBPS &
57 & 24 & 63 & 49 
& 48 & 10
& 56 & 34 & 21 
& 74 & 59 & 36\\
+ BBPS, + pre-optimized &
58 & 24 & 64 & 50 
& 49 & 10 
& 60 & 35 & 22 
& 82 & 65 & 39\\
\bottomrule
\end{tabular}}

\begin{minipage}{\linewidth}\footnotesize
$^\text{a}$Results for PRIMA are obtained by running ERAN (\url{https://github.com/eth-sri/eran}) which contains PRIMA. PRIMA does not support Sine or GeLU activations.\\
$^\text{b}$We have extended its support to GeLU, as discussed in \Cref{ap:gelu}. \\
$^\text{c}$We have extended optimizable linear relaxation in \abcrown to Sine and GeLU, as discussed in \Cref{ap:linear_relaxation_opt}.\\
\end{minipage}

\label{tab:cifar}
\end{table*}

\paragraph{Experiments on feedforward NNs with various activation functions for CIFAR-10.}
In \Cref{tab:cifar}, we show results for models  on CIFAR-10.
On all the models, GenBaB verifies much more instances compared to \abcrown without BaB.
We also conduct ablation studies to investigate the effect of our BBPS heuristic and branching points, with results shown in the last three rows of \Cref{tab:cifar}.
Comparing ``Base BaB'' and `` + BBPS'', on most of the models, we find that our BBPS heuristic significantly improves over directly generalizing the BaBSR heuristic~\citep{bunel2020branch} used in ``Base BaB''.
Comparing ``+ BBPS'' and ``+ BBPS, + pre-optimized'', we find that our pre-optimized branching points achieve a noticeable improvement on many models over always branching in the middle.
The results demonstrate the effectiveness of our GenBaB with our BBPS heuristic and  pre-optimized branching points. 
GenBaB also exhibits much better scalability, where we compare the model size each method can handle w.r.t. a threshold on the number of verified instances. For example, if our threshold is 20 verified instances, GenBaB can at least scale to $4\times 500$ (22 instances verified) while \abcrown w/o BaB cannot even scale to $4\times 100$ (likely even much smaller, as only 4 instances are verified for $4\times 100$).

For PRIMA and vanilla CROWN, as we only use relatively hard instances for verification here, these two methods are unable to verify any instance in this experiment.
For VeriNet, all the models here are too large without a license for the FICO Xpress solver (an academic license was not available to us as mentioned in \Cref{tab:mnist_eran}); we have not obtained the code to run \citet{wu2022toward} on these models. Thus, we do not include the results for VeriNet or \citet{wu2022toward}.

\begin{table}[t]
\centering
\caption{Number of verified instances out of 100 instances on LSTMs and ViTs.
The MNIST model is from PROVER~\citep{ryou2021scalable} with $\eps=0.01$, and the CIFAR-10 models are trained by ourselves with $\eps=1/255$.
``LSTM-7-32'' indicates an LSTM with 7 input frames and 32 hidden neurons, similar for the other two models. ``ViT-$L$-$H$'' stands for $L$ layers and $H$ heads.
Some models have fewer than 100 instances, after filtering out easy or impossible instances, as shown in ``upper bounds''.
Results for PROVER are obtained by running the tool (\url{https://github.com/eth-sri/prover}). 
Results for DeepT are obtained by running the tool (\url{https://github.com/eth-sri/DeepT}).
PROVER and DeepT specialize in RNNs and ViTs, respectively. 
}
\adjustbox{max width=.9\textwidth}{
\begin{tabular}{lccccccc}
\toprule
\multirow{2}{*}{Method} & MNIST Model & \multicolumn{6}{c}{CIFAR-10 Models}\\
& LSTM-7-32 & LSTM-4-32 & LSTM-4-64 & ViT-1-3 & ViT-1-6 & ViT-2-3 & ViT-2-6\\
\midrule
PROVER & 63 & 8 & 3 & - & - & - & -\\
DeepT & - & - & - & 0 & 1 & 0 & 1 \\
\abcrown w/o BaB & 82 & 16 & 9 & 1 & 3 & 11 & 7\\
\textbf{GenBaB (ours)} & 
\textbf{84} & \textbf{20} & \textbf{14} 
& \textbf{49} & \textbf{72} & \textbf{65} & \textbf{56}\\
\midrule
Upper bound & 98 & 100 & 100 & 67 & 92 & 72 & 69\\
\bottomrule
\end{tabular}}
\label{tab:lstm_vit}
\end{table}

\paragraph{Experiments on LSTMs.}
Next, we experiment on LSTMs containing more complex nonlinearities, including both Sigmoid and Tanh activations, as well as multiplication as $\operatorname{sigmoid}(x)\tanh(y)$ and $\operatorname{sigmoid}(x)y$.
We compare with PROVER~\citep{ryou2021scalable} which is a specialized verifier for RNNs and it outperforms earlier works~\citep{ko2019popqorn}.
While there are other works on verifying RNN and LSTM, such as \cite{du2021certrnn,mohammadinejad2021diffrnn,paulsen2022linsyn}, we have not obtained their code, and we also make orthogonal contributions compared to them on improving the relaxation for RNN verification which can also be combined with our BaB. 
We take the hardest model, an LSTM for MNIST, from the main experiments of PROVER (other models can be verified by PROVER on more than 90\% instances and are thus omitted), where each $28\times 28$ image is sliced into 7 frames for LSTM.
We also have two LSTMs trained by ourselves on CIFAR-10, where we linearly map each $32\times 32$ image into 4 patches as the input tokens, similar to ViTs with patches~\citep{dosovitskiy2020image}.
\Cref{tab:lstm_vit} shows the results.
\abcrown without BaB can already outperform PROVER with specialized relaxation for RNN and LSTM.
Our GenBaB outperforms both PROVER and \abcrown without BaB.

\paragraph{Experiments on ViTs.}
We also experiment on ViTs which contain more other nonlinearities, as shown in \Cref{tab:models}.
For ViTs, we compare with DeepT~\citep{bonaert2021fast} which is specialized for verifying Transformers without BaB. We show the results in \Cref{tab:lstm_vit}, where our methods outperform DeepT, and our GenBaB effectively improves the verification.
Moreover, in \Cref{ap:additional_vit}, we compare with \citet{wei2023convex} which supports verifying attention networks but not the entire ViT, and we experiment on models from \citet{wei2023convex} and find that our GenBaB also outperforms \citet{wei2023convex}.

\paragraph{Experiments on ML4ACOPF.}
Finally, we experiment on models for the Machine Learning for AC Optimal Power Flow (ML4ACOPF) problem~\citep{guha2019machine}, and we adopt the ML4ACOPF neural network verification benchmark, a standardized benchmark in the 2023 International Verification of Neural Networks Competition (VNN-COMP'23). The benchmark consists of a NN with power demands as inputs, and the output of the NN gives an operation plan of electric power plants. Then, the benchmark aims to check for a few nonlinear constraint violations of this plan, such as power generation and balance constraints. These constraints, as part of the computational graph to verify, involve many nonlinearities including Sine, Sigmoid, multiplication, and square function.
Our work is the first to support this verification problem.
Among the 23 benchmark instances, PGD attack finds a counterexample on one instance, and our GenBaB verifies all the remaining 22 instances. Only 16 instances can be verified if BaB is disabled. 
This experiment shows a more practical application of our work and further demonstrates the effectiveness of our framework.

\subsection{Time Cost}
\label{sec:time}

\begin{figure*}[t]
\centering
\includegraphics[width=\textwidth]{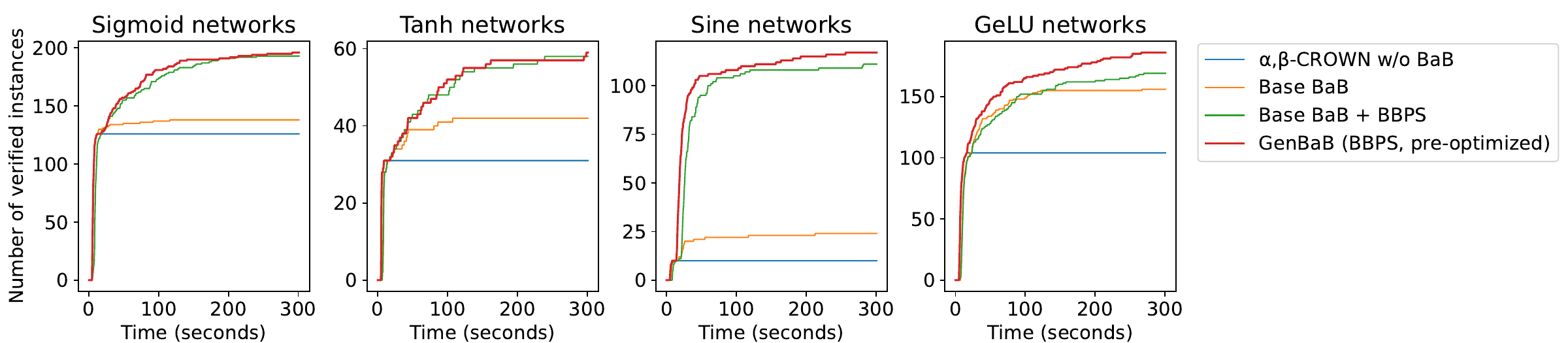}
\caption{Total number of verified instances against running time threshold on feedforward networks for CIFAR-10 with various activation functions.
``Base BaB'' means that in the most basic BaB setting, we use a generalized BaBSR heuristic and always branch in the middle point of intermediate bounds.
``Base + BBPS'' uses our BBPS heuristic. 
Our full GenBaB uses both BBPS and pre-optimized branching points.
}
\label{fig:time}
\end{figure*}

\begin{table}[ht]
\iftoggle{arxiv}{}
{\vspace{-30pt}}
\centering
\caption{
Time cost of pre-optimizing the branching points for models with different nonlinearities.
We only need to run pre-optimization once for each model. The cost is thus negligible as we have many data instances to verify. 
}
\adjustbox{max width=.65\textwidth}{
\begin{tabular}{ccccccc}
\toprule
Model & Sigmoid & Tanh & Sin & GeLU & LSTM & ViT \\
\midrule
Time cost (seconds) & 
49 & 55 & 112 & 82 & 761 & 746 \\
\bottomrule
\end{tabular}}
\label{tab:time_pre_opt}
\vspace{-10pt}
\end{table}

In this section, we analyze the time cost of our method. Our GenBaB aims to verify additional instances which cannot be verified without BaB, for models with general nonlinearities. 
Average time is not a suitable metric here~\citep{wang2021beta}, because different methods verify different numbers of instances, and a stronger verifier which can verify more hard instances requiring more time cost will naturally have a  larger average time compared to a weak verifier which can only verify the easiest instances quickly. 
Instead, we plot the number of verified instances against different time thresholds in \Cref{fig:time}. Such plots, a.k.a. ``cactus plots'', are commonly adopted in previous works~\citep{wang2021beta,brix2023fourth}. The plots show that our GenBaB enables the verification of more instances as more time budget is allowed for BaB.
While the baseline without BaB can verify some relatively easy instances within a short running time (GenBaB can also verify these easy instances during the initial verification with the same time cost if BaB is not needed), the baseline cannot utilize the remaining time budget to verify more instances.
Time cost for LSTM and ViT models are shown in \Cref{ap:time}.
In \Cref{tab:time_pre_opt}, we also show the time cost of pre-optimizing the branching points. Overall, the pre-optimization can be done quickly.
As explained in \Cref{sec:branching_points}, this time cost is negligible for the overall verification, as we only need to run the pre-optimization once for each model and the produced lookup table of branching points can be used to verify an arbitrary number of instances. 

\subsection{Comparison with BaB on ReLU for Models Containing ReLU}

\iftoggle{arxiv}{}{\vspace{-25pt}}
\begin{table}[ht]
\centering
\caption{
Number of verified instances by GenBaB compared to BaB on ReLU only, for certain models containing ReLU.
For BaB on ReLU only, we show results for two different branching heuristic (FSB~\citep{de2021improved} and our BBPS).
}
\adjustbox{max width=.8\textwidth}{
\begin{tabular}{lccccc}
\toprule
Method & ViT-1-3 & ViT-1-6 & ViT-2-3 & ViT-2-6 & ML4ACOPF\\
\midrule
BaB on ReLU only (FSB) & 47 & 70 & 63 & 55 & 18\\
BaB on ReLU only (BBPS) & 47 & 70 & 63 & 55 & 21\\
GenBaB & \textbf{49} & \textbf{72} & \textbf{65} & \textbf{56} & \textbf{22}\\
\hline
Upper bound & 67 & 92 & 72 & 69 & 22\\
\bottomrule
\end{tabular}}
\label{tab:relu_only}
\end{table}

Although our focus is on BaB on non-ReLU nonlinearities, some of the relatively complicated models involved in our experiments still contain ReLU, and thus we compare our GenBaB with BaB on ReLU only for these models. Specifically, only ViT and ML4ACOPF models in our experiments contain ReLU, although they also contain many other nonlinearities.
We show results in \Cref{tab:relu_only}. The results demonstrate that our GenBaB which branches on general nonlinearities outperforms BaB on ReLU only for the models containing ReLU. And many other models with other nonlinearities do not even contain ReLU. Threfore, our GenBaB is important for the BaB on models with general nonlinearities.
We also observe that when we only conduct BaB on ReLU for ML4ACOPF, our BBPS heuristic also outperforms the FSB heuristic~\citep{de2021improved} which is the default branching heuristic adopted by \abcrown for ReLU (FSB is improved from BaBSR~\citep{bunel2020branch} and enhanced with a filtering mechanism to compute actual verified bounds for a shortlist of neurons), and our GenBaB which considers all the nonlinearities can verify more instances (all the 22 possible instances are verified) compared to BaB on ReLU only. 

\section{Related Work}
Due to the NP-complete nature of the NN verification~\citep{katz2017reluplex}, linear bound propagation~\citep{wong2018provable,zhang2018efficient,singh2019abstract} has been proposed to relax nonlinearities in a NN network using linear lower and upper bounds and then propagate the linear relationship between different layers, so that tractable output bounds can be efficiently computed for much larger NNs with various architectures~\citep{boopathy2019cnn,ko2019popqorn,shi2020robustness,xu2020automatic}.
A limitation of using linear bound propagation only is that the linear relaxation, which depends on the output bounds of intermediate layers, can often have a limited tightness as the intermediate bounds gradually become looser in later layers.
Therefore, branch-and-bound (BaB) has been an essential technique in state-of-the-art verifiers~\citep{bunel2018unified,lu2020neural,wang2018efficient,xu2020fast,de2021improved,kouvaros2021towards,wang2021beta,henriksen2021deepsplit,shi2022efficiently,wu2022toward} leveraging linear relaxation, which iteratively branches the intermediate bounds of selected neurons to enable tight linear relaxation and compute tighter output bounds. 
However, most of the existing works on the BaB for NN verification have focused on ReLU networks with the piecewise-linear ReLU activation function, and they are not directly applicable to NNs with nonlinearities beyond ReLU. 
Nevertheless, there are several previous works on the BaB for verifying NNs with nonlinearities other than ReLU.
\citet{henriksen2020efficient} conducted BaB on Sigmoid and Tanh networks, but their framework depends on a commercial LP solver which has been argued as less effective than recent NN verification methods using linear bound propagation~\citep{wang2021beta}.
Besides, \citet{wu2022toward} studied verifying Sigmoid networks with counter-example-guided abstraction refinement.
These works have focused on S-shaped activations such as Sigmoid and Tanh, and there still lacks a general framework supporting general nonlinearities beyond a particular type of activation functions, which we address in this paper.

Orthogonal to our contributions on BaB for general nonlinearities, 
many works studied the verification of NNs with various nonlinearities without considering BaB,
by improving the linear relaxation or extending the support of verification to various architectures or specifications: 
Sigmoid and Tanh networks~\citep{zhang2018efficient,boopathy2019cnn,choi2023reachability}, 
RNNs and LSTMs~\citep{ko2019popqorn,du2021certrnn,ryou2021scalable,mohammadinejad2021diffrnn,zhang2023rnn,tran2023verification},  Transformers~\citep{shi2019robustness,bonaert2021fast,wei2023convex,zhang2024galileo},
general computational graphs~\citep{xu2020automatic},
and specifications on activation patterns instead of input~\citep{geng2023towards}.
Contributions along these lines may be combined with our work, as our BaB is independent from the underlying linear relaxation adopted.
Moreover, some works improved the branching heuristic for verifying ReLU networks:
\citet{lu2020neural} proposed to use a Graph Neural Network for the branching heuristic;
\citet{de2021improved} proposed Filtered Smart Branching (FSB) which filters initial candidates by a heuristic score and then uses a more accurate bound computation to select an optimal neuron from a shortlist;
\citet{ferrari2021complete} considered the effect of a tighter multi-neuron relaxation in the branching heuristic. 
These insights originally for ReLU networks may inspire future improvement of the BaB for general nonlinearities.
\section{Conclusion}
\label{sec:conc}

To conclude, we propose a general BaB framework for NN verification involving general nonlinearities in general computational graphs. We also propose a new branching heuristic for deciding  branched neurons and a pre-optimization procedure for deciding branching points. Experiments on verifying NNs with various nonlinearities demonstrate the effectiveness of our method. 



\section*{Acknowledgments} 
This project is supported in part by NSF 2048280, 2331966, 2331967 and ONR N00014-23-1-2300:P00001. Huan Zhang is supported in part by the AI2050 program at Schmidt Sciences (Grant \#G-23-65921).

\bibliographystyle{iclr2025_conference}
\bibliography{papers}

\newpage
\appendix
\section{Additional Illustration}

\begin{figure}[ht]
\centering
\includegraphics[width=.9\textwidth]{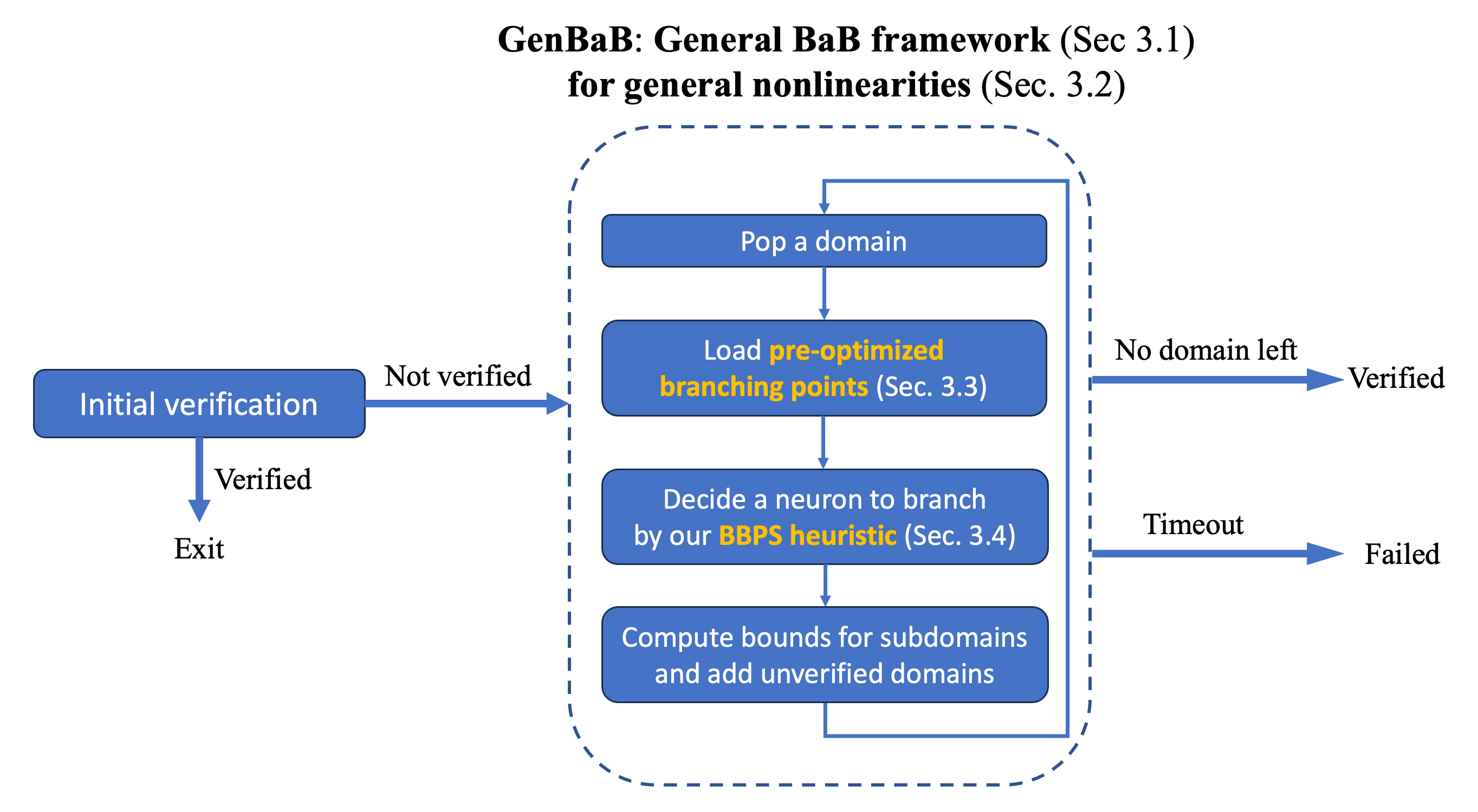}
\caption{
Illustration of our new GenBaB framework summarized in \Cref{sec:framework}.
}
\label{fig:framework}  
\end{figure}

\Cref{fig:framework} illustrates our proposed framework.

\section{Additional Optimizable Linear Relaxation}
\label{ap:linear_relaxation_opt}

In this section, we derive new optimizable linear relaxation for nonlinearities including multiplication, sine, and GeLU, which are not originally supported in \abcrown for optimizable linear relaxation.

\subsection{Optimizable Linear Relaxation for Multiplication}
\label{ap:opt_mul}

For each elementary multiplication $xy$ where $x\in[\ul{x},\ol{x}],\,y\in[\ul{y},\ol{y}]$ are the intermediate bounds for $x$ and $y$, we aim to relax and bound $xy$ as:
\begin{equation}
\forall x\in[\ul{x},\ol{x}], y\in[\ul{y},\ol{y}],
\quad
\ul{a}x+\ul{b}y+\ul{c}\leq xy\leq  \ol{a}x+\ol{b}y+\ol{c},
\label{eq:mul_relaxation}
\end{equation}
where $\ul{a},\ul{b},\ul{c},\ol{a},\ol{b},\ol{c}$ are parameters in the linear bounds.
\citet{shi2019robustness} derived optimal parameters that minimize the gap between the relaxed upper bound and the relaxed lower bound:
\begin{align}
&\argmin_{\ul{a},\ul{b},\ul{c},\ol{a},\ol{b},\ol{c}} \enskip \int_{x\in[\ul{x},\ol{x}]} \int_{y\in[\ul{y},\ol{y}]}
(\ol{a}x+\ol{b}y+\ol{c}) - (\ul{a}x+\ul{b}y+\ul{c})\nonumber\\
&\text{s.t.}\enskip \text{\eqref{eq:mul_relaxation} holds}.
\label{eq:mul_obj}
\end{align}
However, the optimal parameters they found only guarantee that the linear relaxation is optimal for this node, but not the final bounds after conducting a bound propagation on the entire NN.
Therefore, we aim to make these parameters optimizable to tighten the final bounds as previous works did for ReLU networks or S-shaped activations~\citep{xu2020fast,lyu2019fastened}.

We notice that \citet{shi2019robustness} mentioned that there are two solutions for  $\ul{a},\ul{b},\ul{c}$ and $\ol{a},\ol{b},\ol{c}$ respectively that solves \eqref{eq:mul_obj}:
\begin{equation}
\left\{\begin{aligned}
&\ul{a}_1= \ul{y}\\
&\ul{b}_1=\ul{x}\\
&\ul{c}_1=-\ul{x}\ul{y}
\end{aligned}\right.,
\quad\left\{\begin{aligned}
&\ol{a}_1= \ol{y}\\
&\ol{b}_1=\ul{x}\\
&\ol{c}_1=-\ul{x}\ol{y}
\end{aligned}\right.,
\label{eq:mul_sol_1}
\end{equation}
\begin{equation}
\left\{\begin{aligned}
&\ul{a}_2= \ol{y}\\
&\ul{b}_2=\ol{x}\\
&\ul{c}_2=-\ol{x}\ol{y}
\end{aligned}\right.,
\quad\left\{\begin{aligned}
&\ol{a}_2= \ul{y}\\
&\ol{b}_2=\ol{x}\\
&\ol{c}_2=-\ol{x}\ul{y}
\end{aligned}\right..
\label{eq:mul_sol_2}
\end{equation}
Therefore, to make the parameters optimizable, we introduce parameters $ \ul{\alpha} $ and $\ol{\alpha}$,  and we interpolate between \eqref{eq:mul_sol_1} and \eqref{eq:mul_sol_2} as:
\begin{equation}
\left\{\begin{aligned}
&\ul{a}= \ul{\alpha} \ul{y} + (1-\ul{\alpha}) \ol{y}\\
&\ul{b}= \ul{\alpha} \ul{x} + (1-\ul{\alpha})\ol{x}\\
&\ul{c}=-\ul{\alpha}\ul{x}\ul{y} - (1-\ul{\alpha})\ol{x}\ol{y}
\end{aligned}\right. \quad \text{s.t.} \enskip 0\leq\ul{\alpha}\leq 1,
\end{equation}
\begin{equation}
\left\{\begin{aligned}
&\ol{a}= \ol{\alpha} \ol{y} + (1-\ol{\alpha}) \ul{y}\\
&\ol{b}= \ol{\alpha} \ul{x} + (1-\ol{\alpha})\ol{x}\\
&\ol{c}=-\ol{\alpha}\ul{x}\ol{y} - (1-\ol{\alpha})\ol{x}\ul{y}
\end{aligned}\right. \quad \text{s.t.} \enskip 0\leq\ol{\alpha}\leq 1.
\end{equation}
It is easy to verify that interpolating between two sound linear relaxations satisfying \eqref{eq:mul_relaxation} still yields a sound linear relaxation.
And $\ul{\alpha}$ and $\ol{\alpha}$ are part of all the optimizable linear relaxation parameters $\rvalpha$ mentioned in \Cref{sec:bg}.

\subsection{Optimizable Linear Relaxation for Sine}

We also derive new optimized linear relaxation for periodic functions, in particular $\sin(x)$.
For $sin(x)$ where $x\in[\ul{x},\ol{x}]$, we aim to relax and bound $sin(x)$ as:
\begin{equation}
\forall x\in[\ul{x},\ol{x}],
\quad
\ul{a}x+\ul{b}\leq \sin(x)\leq  \ol{a}x+\ol{b},
\label{eq:sin_relaxation}
\end{equation}
where $\ul{a},\ul{b}, \ol{a},\ol{b}$ are parameters in the linear bounds.
A non-optimizable linear relaxation for $\sin$ already exists in \abcrown and we adopt it as an initialization and focus on making it optimizable.
At initialization, we first check the line connecting $(\ul{x}, \sin(\ul{x}))$ and $(\ol{x}, \sin(\ol{x}))$, and this line is adopted as the lower bound or the upper bound without further optimization, if it is a sound bounding line.

\begin{figure}[ht]
\centering
\includegraphics[width=.6\textwidth]{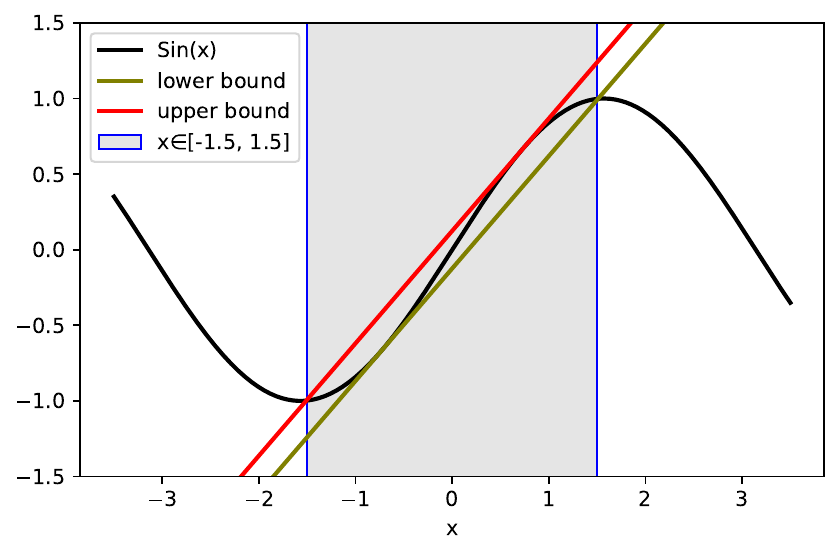}
\caption{Linear relaxation for a Sin activation in an input range $[-1.5,1.5]$ where the function is S-shaped.}
\label{fig:sin_sigmoid_like}
\end{figure}

Otherwise, a tangent line is used as the bounding line with the tangent point being optimized.
Within $[\ul{x}, \ol{x}]$, if $\sin(x)$ happens to be monotonic with at most only one inflection point, the tangent point can be optimized in a way similar to bounding an S-shaped activation~\citep{lyu2019fastened}, as illustrated in \Cref{fig:sin_sigmoid_like}.

\begin{figure*}[ht]
\centering
\begin{subfigure}[b]{0.48\textwidth}
\centering
\includegraphics[width=\textwidth]{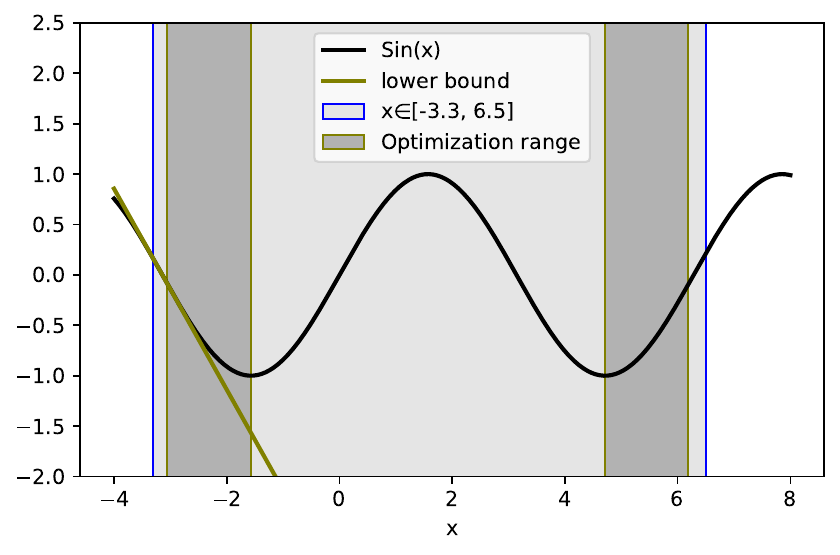}
\caption{The lower bound of Sin activation when $\alpha_l = \ul{\alpha}_l$.}
\label{fig:sin_opt_left}
\end{subfigure}
\hfill
\begin{subfigure}[b]{0.48\textwidth}
\centering
\includegraphics[width=\textwidth]{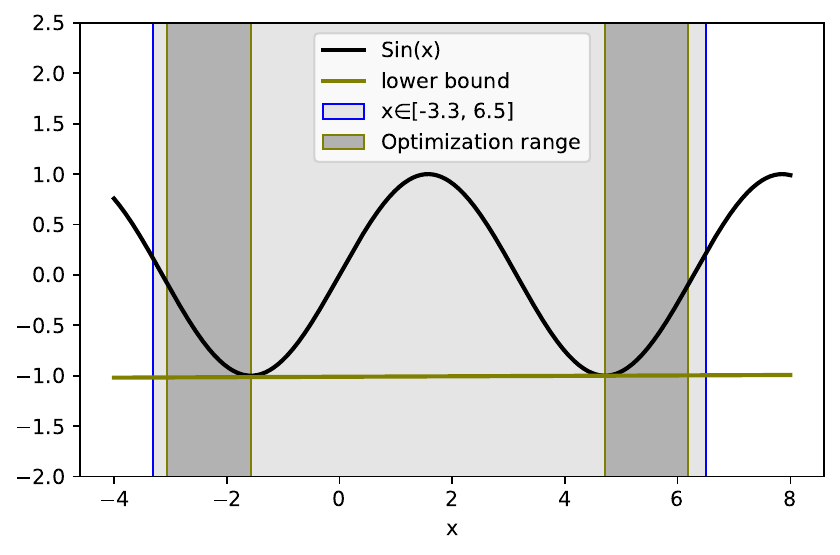}
\caption{The lower bound of Sin activation when $\alpha_l = 1.5\pi$.}
\label{fig:sin_opt_middle}
\end{subfigure}\\
\begin{subfigure}[b]{0.48\textwidth}
\centering
\includegraphics[width=\textwidth]{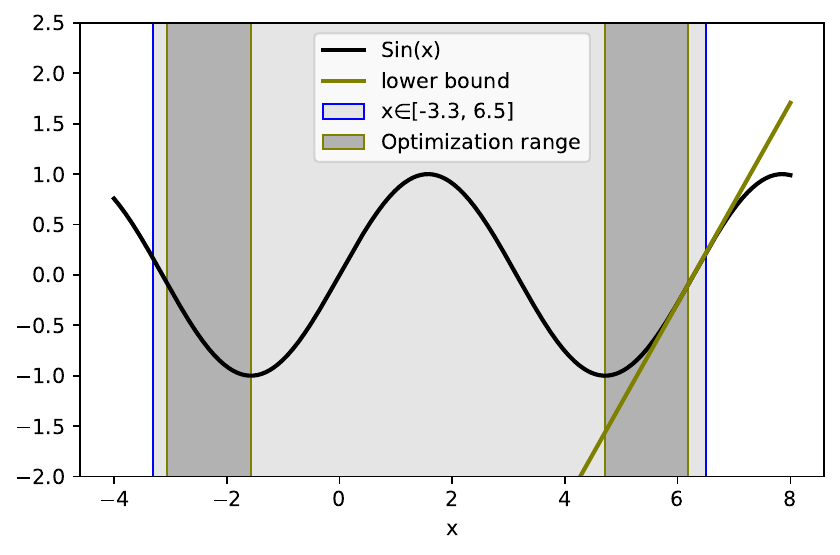}
\caption{The lower bound of Sin activation when $\alpha_l = \ol{\alpha}_l $.}
\label{fig:sin_opt_right}
\end{subfigure}
\caption{Optimizing the lower bound of a Sin activation, where ``Optimization range'' shows all the valid tangent points for the lower bound during the optimization.}
\label{fig:sin}
\end{figure*}

Otherwise, there are multiple extreme points within the input range. Initially, we aim to find a tangent line that passes $(\ul{x}, \sin(\ul{x}))$ as the bounding line.
Since $\ul{x}$ may be at different cycles of the $\sin$ function, we project into the cycle with range $[-0.5\pi, 1.5\pi]$, by taking $\ul{\tilde{x}}_l = \ul{x} - 2\ul{k}_l\pi$, where $\ul{k}_l = \floor{\frac{\ul{x} + 0.5\pi}{2\pi}}$.
With a binary search, we find a tangent point $\ul{\alpha}_l$ on the projected cycle that satisfies
\begin{equation}
 \sin'(\ul{\alpha}_l)(\ul{\alpha}_l - \ul{\tilde{x}}_l) + \sin(\ul{\tilde{x}}_l)= \sin(\ul{\alpha}_l),\\
\end{equation}
which corresponds to a tangent point $\ul{t}_l = \ul{\alpha}_l+2k_l\pi $ at the original cycle of $\ul{x}$, and for any tangent point within the range of $[\ul{\alpha}_l + 2\ul{k}_l\pi, 1.5\pi + 2\ul{k}_l\pi]$, the tangent line is a valid lower bound.
Similarly, we also consider the tangent line passing $(\ol{x}, \sin(\ol{x}))$, and we take
$\tilde{\ol{x}}_l = \ol{x} - 2\ol{k}_l\pi$,
where $\ol{k}_l= \floor{\frac{\ol{x} - 1.5\pi}{2\pi}}$,
so that $\tilde{\ol{x}}_l$ is within range $[1.5\pi, 3.5\pi]$.
We also conduct a binary search to find the tangent point $\ol{\alpha}_l$, which corresponds to to $ \ol{\alpha}_l+2\ol{k}_l\pi$ in the original cycle of $\ol{x}$, and for any tangent point within the range $[1.5\pi+ 2\ol{k}_l\pi, \ol{\alpha}_l + 2\ol{k}_l\pi]$, the tangent line is also a valid lower bound.
We make the tangent point optimizable with a parameter $\alpha_l~(\ul{\alpha_l}\leq\alpha_l\leq\ol{\alpha}_l)$, which corresponds to a tangent line at tangent point $t_l$ as the lower bound in \eqref{eq:sin_relaxation} and \Cref{fig:sin}:
\begin{align}
&\left\{\begin{aligned}
&\ul{a}= \sin'(t_l)\\
&\ul{b}= \sin(t_l) - \ul{a}t_l\\
\end{aligned}\right.,\\
&\text{where}\enskip
\left\{\begin{aligned}
&t_l = \alpha_l + 2\ul{k}_l\pi \enskip \text{if} \enskip\ul{\alpha}_l \leq \alpha_l \leq 1.5\pi\\
&t_l = \alpha_l + 2\ol{k}_l\pi \enskip \text{if}\enskip1.5\pi < \alpha_l \leq \ol{\alpha}_l \\
\end{aligned}\right..
\end{align}
In particular, when $\alpha_l=1.5\pi$, both $\alpha_l+2\ul{k}_l\pi$ and $\alpha_l+2\ol{k}_l\pi$ are tangent points for the same tangent line.

The derivation for the upper bound is similar.
We take $\ul{\tilde{x}}_u = \ul{x} - 2\ul{k}_u\pi$, where $\ul{k}_u= \floor{\frac{\ul{x} - 0.5\pi}{2\pi}}$, so that $\ul{\tilde{x}}_u$ is in range $[0.5\pi, 2.5\pi]$.
And we take  $\tilde{\ol{x}}_u = \ol{x} - 2\ol{k}_u\pi$, where $\ol{k}_u= \floor{\frac{\ol{x} - 2.5\pi}{2\pi}}$, so that $\tilde{\ol{x}}_u$ is in range $[2.5\pi, 4.5\pi]$.
Let $\ul{\alpha}_u$ be the tangent point where the tangent line crosses $\ul{\tilde{x}}_u$,
and $\ol{\alpha}_u$ be the tangent point where the tangent line crosses $\tilde{\ol{x}}_u$, as found by a binary search.
We define an optimizable parameter
$\alpha_u~(\ul{\alpha_u}\leq\alpha_u\leq\ol{\alpha}_u)$ which corresponds to a tangent line as the upper bound:
\begin{align}
&\left\{\begin{aligned}
&\ol{a}= \sin'(t_u)\\
&\ol{b}= \sin(t_u) - \ol{a}t_u\\
\end{aligned}\right.,\\
&\text{where}\enskip
\left\{\begin{aligned}
&t_u = \alpha_u + 2\ul{k}_u\pi \enskip\text{if}\enskip \ul{\alpha}_u \leq \alpha_u \leq 2.5\pi \\
&t_u = \alpha_u + 2\ol{k}_u\pi \enskip\text{if}\enskip 2.5\pi < \alpha_u \leq \ol{\alpha}_u\\
\end{aligned}
\right. .
\end{align}

\subsection{Optimizable Linear Relaxation for GeLU}

\label{ap:gelu}

For GeLU function where $x\in[\ul{x},\ol{x}]$ are the intermediate bounds for $x$, we aim to relax and bound $\operatorname{GeLU}(x)$ as:
\begin{equation}
\forall x\in[\ul{x},\ol{x}],
\quad
\ul{a}x+\ul{b}\leq \operatorname{GeLU}(x)\leq  \ol{a}x+\ol{b},
\label{eq:gelu_relaxation}
\end{equation}
where $\ul{a},\ul{b}, \ol{a},\ol{b}$ are parameters in the linear bounds.

Given input range $[\ul{x}, \ol{x}]$, if $\ol{x} \leq 0$ or $\ul{x} \geq 0$, the range contains only one inflection point, the tangent point can be optimized in a way similar to bounding an S-shaped activation~\citep{lyu2019fastened}.
In other cases, $\ul{x} < 0$ and $\ol{x} > 0$ holds. For the upper bound, we use the line passing $(\ul{x},\operatorname{GeLU}(\ul{x}))$ and $(\ol{x},\operatorname{GeLU}(\ol{x}))$.
For the lower bound, we derive two sets of tangent lines that crosses $(\ul{x}, \operatorname{GeLU}(\ul{x}))$ and $(\ol{x}, \operatorname{GeLU}(\ol{x}))$ with tangent points denoted as $\ul{\alpha}$ and $\ol{\alpha}$ respectively. We determine $\ul{\alpha}, \ol{\alpha}$ using a binary search that solves:
\begin{equation}
\left\{\begin{aligned}
&\operatorname{GeLU}'(\ul{\alpha})(\ul{\alpha} - \ul{x}) + \operatorname{GeLU}(\ul{x})= \operatorname{GeLU}(\ul{\alpha})\\
&\operatorname{GeLU}'(\ol{\alpha})(\ol{\alpha} - \ol{x}) + \operatorname{GeLU}(\ol{x})= \operatorname{GeLU}(\ol{\alpha})\\
\end{aligned}\right..
\end{equation}
Any tangent line with a tangent point $\alpha~(\ul{\alpha}\leq\alpha\leq\ol{\alpha})$ is a valid lower bound, which corresponds to the lower bound in \eqref{eq:gelu_relaxation} with:
\begin{equation}
\left\{\begin{aligned}
&\ul{a}= \operatorname{GeLU}'(\alpha)\\
&\ul{b}= \operatorname{GeLU}(\alpha) - \alpha \operatorname{GeLU}'(\alpha)\\
\end{aligned}\right. \quad \text{s.t.} \enskip \ul{\alpha}\leq \alpha \leq \ol{\alpha}.
\end{equation}

\section{Additional Results}
\label{ap:additional_results}

\subsection{Time Cost on LSTM and ViT}
\label{ap:time}
\begin{figure*}[t]
\centering
\includegraphics[width=.85\textwidth]{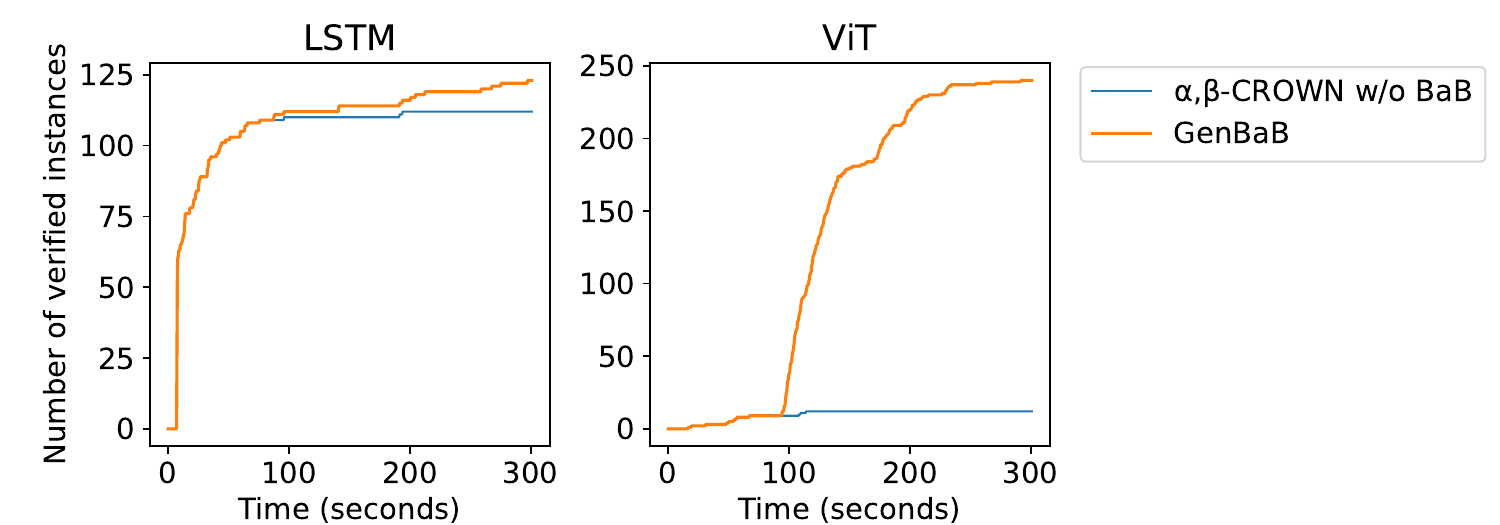}
\caption{Total number of verified instances against running time threshold on LSTM and ViT.
}
\label{fig:time_lstm_vit}
\end{figure*}

In \Cref{fig:time_lstm_vit}, we show the time cost on LSTM and ViT models, as discussed in \Cref{sec:time}.

\subsection{Experiments on Self-Attention Networks from \cite{wei2023convex}}
\label{ap:additional_vit}

To compare with \citet{wei2023convex} which only supports verifying single-layer self-attention networks but not the entire ViT, we adopt pre-trained models from \citet{wei2023convex} and run our verification methods under their settings, with 500 test images in MNIST using $\eps=0.02$. We show the results in \Cref{tab:vit&marabou}, where our methods also outperform \citet{wei2023convex} on all the models.

\begin{table}[ht]
\centering
\caption{Number of verified instances out of 500 instances in MNIST with $\eps=0.02$. A-small, A-medium and A-big are three self-attention networks with different parameter sizes from \citet{wei2023convex}.
}
\adjustbox{max width=.8\textwidth}{
\begin{tabular}{lccc}
\toprule
Method & A-small & A-medium & A-big\\
\midrule
\citet{wei2023convex} & 406 & 358 & 206 \\
\hline
\abcrown w/o BaB & 444 & 388 & 176\\
\textbf{GenBaB (ours)} & \textbf{450} & \textbf{455} & \textbf{232}\\
\hline
Upper bound & 463 & 479 & 482\\
\bottomrule
\end{tabular}}
\label{tab:vit&marabou}
\end{table}

\subsection{Experiments on a ResNet Model}
\label{ap:resnet}

In this section, we demonstrate our method on a ResNet model~\citep{he2016deep}.
The model has the same size as the one used in \citet{wang2021beta},
which has 2 residual blocks with 6 convolutional layers and fully-connected layers in total. 
Since our focus is on general nonlinearities, we use GeLU activation instead of ReLU. 
We train the model on CIFAR-10 with PGD adversarial training using $\eps=2/255$. 
As the results shown in \Cref{tab:resnet}, our GenBaB significantly improves the verification on the ResNet model compared to \abcrown without BaB.

\begin{table}[ht]
\centering
\caption{Number of verified instances on a ResNet model with the same architecture as the ResNet in \citet{wang2021beta}, but the activation function is GeLU instead of ReLU.
}
\adjustbox{max width=.8\textwidth}{
\begin{tabular}{lccc}
\toprule
Method & ResNet\\
\midrule
\abcrown w/o BaB & 24\\
\textbf{GenBaB (ours)} & \textbf{74}\\
\bottomrule
\end{tabular}}
\label{tab:resnet}
\end{table}

\subsection{Experiments on Larger $\eps$}
\label{ap:large_eps}

In this section, we demonstrate GenBaB on larger $\eps$. 
We consider $\eps=2/255$ and $\eps=8/255$ for $4\times 100$ feedforward networks with various activation functions on CIFAR-10. As the results shown in \Cref{tab:large_eps}, our GenBaB effectively improves the verification on all these models.

\begin{table}[ht]
\centering
\caption{Number of verified instances on $4\times 100$ feedforward networks with various activation functions on CIFAR-10 when a larger $\eps=2/255$ or $\eps=8/255$ is used.
}
\adjustbox{max width=\textwidth}{
\begin{tabular}{lcccccccc}
\toprule
\multirow{2}{*}{Method} & \multicolumn{4}{c}{$\eps=2/255$} & \multicolumn{2}{c}{$\eps=8/255$}\\
& Sigmoid & Tanh & Sin & GeLU & Sigmoid & Tanh & Sin & GeLU \\
\midrule
\abcrown w/o BaB & 
33 & 15 & 11 & 39
& 16 & 11 & 2 & 34\\
\textbf{GenBaB (ours)} & 
\textbf{56} & \textbf{26} & \textbf{65} & \textbf{65} 
& \textbf{37} & \textbf{19} & \textbf{22} & \textbf{35} \\
\bottomrule
\end{tabular}}
\label{tab:large_eps}
\end{table}

\subsection{Experiments on a ReLU Network}
\label{ap:relu}

\begin{table}[ht]
\centering
\caption{
Results on a ``ConvSmall'' model with ReLU activation~\citep{singh2019beyond,singh2019abstract,muller2022prima} on 1000 instances from CIFAR-10. Percentage of instances verified by various methods are reported. For methods other than PRIMA, we use \abcrown as the underlying verifier but vary the branching heuristic. See explanation about the backup score in \Cref{ap:relu}.
}
\adjustbox{max width=.5\textwidth}{
\begin{tabular}{lc}
\toprule
Method & Verified \\
\midrule
PRIMA & 44.6\%\\
BaBSR w/o backup score & 45.6\%\\
BaBSR w/ backup score & 46.2\%\\
Backup score only & 45.0\%\\
BBPS w/o backup score & 46.0\%\\
BBPS w/ backup score & 46.2\%\\
\bottomrule
\end{tabular}}
\label{tab:relu_convsmall}
\end{table}

In this section, we study the effect of our BBPS heuristic on ReLU models. We adopt settings in \citet{singh2019beyond,singh2019abstract,muller2022prima} and experiment on a ``ConvSmall'' model with ReLU activation. The verification is evaluated on 1000 instances on CIFAR-10, following prior works. We show the results in \Cref{tab:relu_convsmall},
We find that on this ReLU network, our BBPS also works better than the BaBSR heuristic, when there is no \emph{backup score} (46.0\% verified by BBPS v.s. 45.6\% verified by BaBSR).
However, we find that recent works typically add a \emph{backup score} for BaBSR, which is another heuristic score that serves as a backup for neurons with extremely small BaBSR scores. The backup score did not exist in the original BaBSR heuristic~\citep{bunel2020branch} but it appeared in \citet{de2021improved} and has also been adopted by works such as \citet{wang2021beta} when using BaBSR for ReLU networks.
This backup score basically uses the intercept of the linear relaxation for the upper bound of a ReLU neuron that needs branching. Unlike BaBSR or BBPS, the backup score does not aim to directly estimate the change on the bounds computed by bound propagation, but aims to use the intercept to reflect the reduction of the linear relaxation after the branching.
When the backup score is combined with BaBSR or BBPS for ReLU networks, the backup score seems to dominate the performance, where both BaBSR and BBPS have similar performance with the backup score added (46.2\% verified), which hides the underlying improvement of BBPS over BaBSR by providing a more precise estimation. However, the backup score is specifically for ReLU, assuming that the intercept of the linear relaxation can reflect the reduction of the linear relaxation, which is not the case for general nonlinearities.
We leave it for future work to study the possibility of designing a backup score for general nonlinearities.

\section{Experimental Details}
\label{ap:details}
\label{ap:exp_details}

\paragraph{Implementation details.}
We implement our GenBaB algorithm based on \abcr\footnote{\url{https://github.com/Verified-Intelligence/alpha-beta-CROWN}} which originally did not support BaB for nonlinearities other than ReLU.
To pre-optimize the branching points, we enumerate the branching points ($p$ in \eqref{eq:opt_bp}) with a step size instead of performing gradient descent, considering that we only have up to two parameters for the branching points in our experiments. For nonlinearities with a single input, we use a step size of 0.01, and for nonlinearities with two inputs, we use a step size of 0.1. We pre-optimize the branching points for intermediate bounds within the range of $[-5, 5]$. For all the experiments, each experiment is run using a single NVIDIA GTX 1080Ti GPU.

\paragraph{Details on training the models.}
To train our models on CIFAR-10, we use PGD adversarial training~\citep{madry2017towards}. We use 7 PGD steps during the training and the step size is set to $\eps/4$.
For training the Sigmoid networks in \Cref{tab:cifar}, we use the SGD optimizer with a learning rate of $5\times 10^{-2}$ for 100 epochs;
and for training the Tanh networks, we use the SGD optimizer with a learning rate of $1\times 10^{-2}$ for 100 epochs.
For training the LSTMs in \Cref{tab:lstm_vit}, we use the Adam optimizer with a learning of $10^{-3}$ for 30 epochs.
And for training the ViTs, we use the Adam optimizer with a learning of $5\times 10^{-3}$ for 100 epochs. For Sine networks, we use the SGD optimizer with a learning rate of $1\times 10^{-3}$ for 100 epochs

\paragraph{Memory cost.}
Memory cost of our framework is highly manageable. 
To store bounds for branched domains, note that the pool of domains with branched intermediate bounds is stored on CPU memory (not GPU), and in each iteration of BaB, only a batch of domains is loaded to GPU and handled in parallel on GPU, the batch size can be configurable to fit the GPU memory (mentioned in \Cref{sec:framework}).
For bound computation during BaB, as mentioned in \Cref{sec:heuristic}, since intermediate bounds are not re-computed during BaB, the space complexity for each subproblem is the same as a regular NN forward pass.

For the lookup table for pre-optimized branching points, the memory cost is also small: 4MB for 1D nonlinearities (tanh, sigmoid, sin, etc.) and 800MB for 2D nonlinearities (multiplication). Although the memory cost of a lookup table  can become larger for higher-dimensional nonlinearities, the granularity of lookup tables is configurable to reduce the memory requirement at the cost of slightly less optimal branching points. For further scalability, future works may compress the lookup table by an NN (such as \citet{julian2016policy}) for high dimensional nonlinearities, and the validity of branching points predicted by NN is easy to guarantee by clipping the prediction.

\end{document}